\documentclass[runningheads]{llncs}
\usepackage{graphicx}
\usepackage{pdfpages}
\usepackage{amsmath,amssymb} % define this before the line numbering.
\usepackage{lineno}
\usepackage{color}
\usepackage{url}

\begin{document}

%macro for raising the point in decimal numbers; see example in the abstract
\newcommand{\point}{
    \raise0.7ex\hbox{.}
    }

%Do   -- NOT --    use any additional macros

\pagestyle{headings}

\mainmatter

%===========================================================
\title{Robust Visual Tracking Using Dynamic Classifier Selection with Sparse Representation of Label Noise}

\titlerunning{Robust Visual Tracking Using Dynamic Classifier Selection} % Replace with your title

\authorrunning{Y. Chen and Q. Wang} % Replace with your names

\author{Yuefeng Chen and Qing Wang} % Replace with your names
\institute{School of Computer Science and Engineering \\Northwestern Polytechnical University, Xi'an 710072, P. R. China} % Replace with your institute's address

\maketitle

%===========================================================
\begin{abstract}
Recently a category of tracking methods based on ``tracking-by-detection" is widely used in visual tracking problem. Most of these methods update the classifier online using the samples generated by the tracker to handle the appearance changes. However, the self-updating scheme makes these methods suffer from drifting problem because of the incorrect labels of weak classifiers in training samples. In this paper, we split the class labels into true labels and noise labels and model them by sparse representation. A novel dynamic classifier selection method, robust to noisy training data, is proposed. Moreover, we apply the proposed classifier selection algorithm to visual tracking by integrating a part based online boosting framework. We have evaluated our proposed method on 12 challenging sequences involving severe occlusions, significant illumination changes and large pose variations. Both the qualitative and quantitative evaluations demonstrate that our approach tracks objects accurately and robustly and outperforms state-of-the-art trackers.
\end{abstract}

%===========================================================
\section{Introduction}

Visual object tracking has been one of the most attractive topics in computer vision and there are numerous practical applications so far, such as motion analysis, video surveillance, traffic controlling and so on. In recent years, although many tracking methods \cite{01,02,03,04,05,06,07,08} have been proposed and made a certain breakthrough, it is still a challenging issue to design a tracking algorithm which is robust to severe occlusions, significant illumination changes, pose variations, fast motion, scale changes and background clutter.

Generally, a typical tracking system contains three basic components: object representation, appearance model and motion model. The appearance model is used to represent the object and provide prediction accordingly. The motion model (such as Kalman filter \cite{09}, Particle filter \cite{02,03,10}) is applied to predict the motion distribution of the object between two adjacent frames. In this paper, we focus on the appearance model which is the most important and challenging part of the tracker.

In literatures \cite{01,02,18,19}, most of tracking approaches can be categorized as either generative or discriminative ones based on different appearance models. The generative methods formulate the tracking problem as locating the object region with {\em maximum} probability generated from the \textbf{modeled} appearance model \cite{03,09}. Adam {\it et al.} \cite{11} proposed a robust fragments-based tracking approach where the object is represented by multiple image fragments. This tracker can handle partial occlusions and pose changes well. Since the template was fixed and not updated during tracking, the tracker is sensitive to appearance changes of the object. To acclimate to appearance changes, the template or appearance model of the tracker should be updated dynamically. Ross {\it et al.} \cite{03} developed a tracking algorithm based on incrementally learning a low-dimensional subspace representation. Kwon {\it et al.} \cite{06} decomposed the observation model into multiple basic observation models and the proposed method was robust to significant motion and appearance changes.

For discriminative methods, the tracking problem is formulated as training a discriminative classifier to separate the object from surrounding background. These methods are also noted as tracking-by-detection methods since training a classifier is similar to detection problem. Recently, numerous tracking approaches based on detection are proposed \cite{01,12,13,14,15,16,17}. In order to handle appearance changes, online ensemble learning was popularly applied to the tracking problem. In \cite{18}, Avidan proposed an online ensemble tracking algorithm by online learning. Grabner {\it et al.} \cite{12} proposed an online Adaboost feature selection method. Babenko {\it et al.} \cite{01} used online MIL (Multiple Instance Learning) instead of traditional supervised learning to handle the drifting problem. In their work, the training samples are constructed as bags and labels are corresponded to bags rather than samples. Classifier selection scheme plays an important role in online ensemble learning based tracking methods. In \cite{12}, classifiers are selected by measuring the accumulated error. In \cite{01}, by maximizing the log likelihood of bags, classifiers are selected from the classifier pool. However, these classifier selection algorithms almost assume the training dataset is clean, which do not meet practical applications of visual tracking. Noisy training dataset will reduce the performance of the classifier and cause drifting during tracking.

Recently, sparse representation \cite{19} has been widely applied in object tracking \cite{02,05,20,21,22,23}. Mei {\it et al.} \cite{02} treated tracking as a sparse approximation problem in a particle filter framework. Each target candidate is sparsely represented in the space of object templates and trivial templates, and the candidate with the smallest projection error is taken as the target. This representation is robust to partial occlusions and other changes. However it is computationally expensive to obtain sparse coefficients by $\ell_1$ {\it minimization}. This restriction makes it unsuitable for real-time applications. Bao {\it et al.} \cite{24} developed a very fast numerical solver to tackle the $\ell_1$ {\it minimization} problem based on the accelerated proximal gradient (APG) algorithm such that the tracker can work in real time. Zhang {\it et al.} \cite{05} expanded the $\ell_1$ tracker by employing popular sparsity-inducing \(\ell_{p,q}\) mixed norms and formulated object tracking in a particle filter framework as a multi-task sparse learning problem. Most of these methods use sparse coding to represent the appearance model of the object. However, little work has been done on sparse representation for the class labels of classifiers.

\begin{figure}[tbp]
\centering
\includegraphics[width=0.95\textwidth]{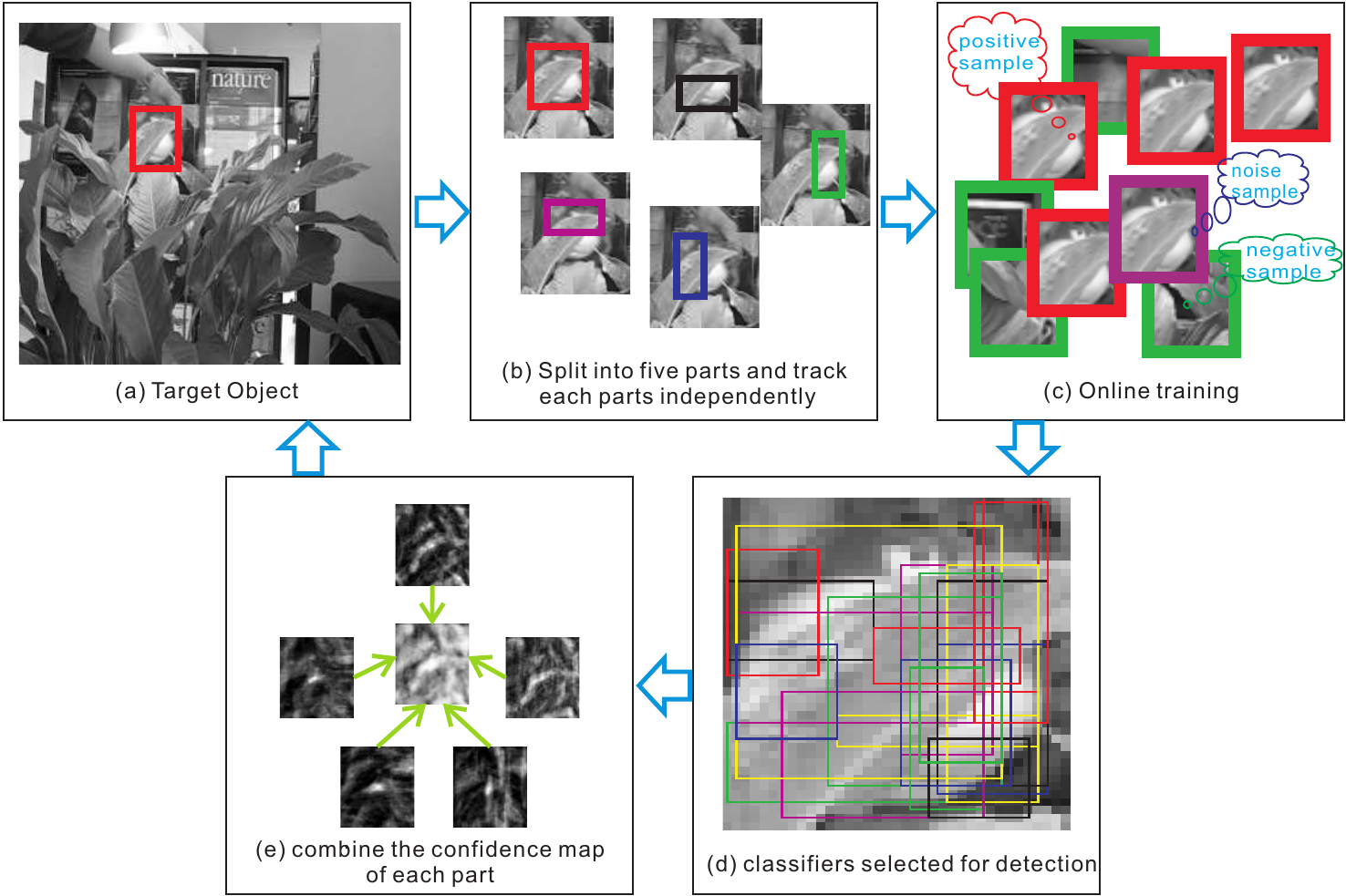}
\caption{The pipeline of the proposed object tracking approach. In (c), the sample with purple box is a noisy one since it is actually positive but labeled as negative.}
\label{fig:pipeline}
\end{figure}

In this paper, we present a novel dynamic classifier selection with sparse representation of label noise. We use sparse representation to model the class labels and obtain classifiers by $\ell_1$ minimization. In order to verify the robustness of the selection scheme, the algorithm is integrated into a boosting framework to solve visual tracking problem. In order to handle severe occlusion issue, we expand the traditional boosting algorithm to a part based one which is more robust to occlusion.  Figure~\ref{fig:pipeline} depicts an overview of the proposed tracking approach. The key contributions of our work can be concluded as follows.

\begin{itemize}
\renewcommand{\labelitemi}{$\bullet$}
  \item We split the class labels into true labels and noisy labels and model them using sparse representation. Moreover, we propose a novel dynamic classifier selection algorithm based on $\ell_1$ {\it minimization}, which is more robust than traditional selection schemes in \cite{01,12}.
  \item We expand online boosting framework and employ the proposed dynamic classifier selection to this framework. And then a more robust tracking approach is proposed and achieves better performance comparing with other methods.
\end{itemize}

The rest of the paper is organized as follows. Dynamic classifier selection is proposed in section 2. In section 3, we integrate dynamic classifier selection and an expanded online boosting framework to object tracking. Experimental results are shown and analyzed in section 4 with qualitative and quantitative evaluations. The concluding remarks are drawn in section 5.

\section{Dynamic classifier selection with label noise}
In this section, we describe the proposed dynamic classifier selection algorithm with class label noise in details. First we discuss the motivation of our work. Then, the model of classifier with class label noise is presented. At last, we solve this problem using {$\ell_1$} minimization.

\subsection{Motivation}
In the field of machine learning, ensemble learning methods has been popular in recent years and achieved better performance than traditional learning methods. One of the most critical tasks is to select a group of classifier dynamically in the weak classifier pool \cite{30}. Many dynamic classifier selection methods \cite{12,01} have been proposed in recent years, however, most of them do not pay attention to the class label noise which is inherent in the real world dataset. The performance of ensemble learning methods can be rapidly degraded when there are some noises in the training dataset. Therefore, developing a noise-tolerant dynamic classifier selection algorithm is a significant task. In this work, we model the class labels using sparse representation based on the assumption that noise class labels are sparsely distributed over the training dataset. Finally, the weak classifier with the maximum value in the sparse vector will be selected.

\subsection{Sparse representation of class labels}
Given the training dataset $ \mathbf{X}= \left [ \mathbf{x}_1, \mathbf{x}_2, \ldots, \mathbf{x}_L  \right ]$ and label  $\mathbf{y} = \left[ y_1, y_2, \ldots, y_L \right]^\top  \in \mathbb{R}^{L \times 1}, y_i \in \left \{ -1, +1 \right \}$. The class label vector predicted by the weak classifier $\mathbf{h}$ in this feature set $\mathbf{X}$ is denoteds as,
\begin{equation}\label{equ:equ7}
    \mathbf{h} \left (  \mathbf{X} \right ) = {
    \left [
    h \left(  \mathbf{x}_1  \right ),
    h \left(  \mathbf{x}_2  \right ),
    \ldots,
    h \left(  \mathbf{x}_L  \right )
    \right]}^\top,
\end{equation}
where $\mathbf{h} \left (  \mathbf{X} \right ) \in \mathbb{R}^{L \times 1}$ and $h \left(  \mathbf{x}_i  \right )$ is the predicted class label of sample $\mathbf{x}_i$ by the weak classifier $h$. Then the joint class labels corresponding to different weak classifiers are denoted by,
\begin{equation}\label{equ:equ8}
    \mathbf{\Phi} =
                \left [
                {\mathbf{h}}_1 \left (  \mathbf{X} \right ),
                {\mathbf{h}}_2 \left (  \mathbf{X} \right ),
                \ldots,
                {\mathbf{h}}_M \left (  \mathbf{X} \right )
            \right ],
\end{equation}
where $ \mathbf{\Phi} \in \mathbb{R}^{L \times M}, L \ll M $, $L$ is the number of training samples, and $M$ is the number of weak classifiers in a pool which are used for selection. Assuming that the true label vector $\hat{\mathbf{y}}$ without noise is a linear combination of ${\mathbf{h}}_i \left (  \mathbf{X} \right )$, thus $\hat{\mathbf{y}}$  can be represented as,
\begin{equation}\label{equ:equ9}
    \hat{\mathbf{y}} \approx \mathbf{\Phi} \pmb{\beta} =
    {\beta}_1 {\mathbf{h}}_1 \left (  \mathbf{X} \right ) +
    {\beta}_2 {\mathbf{h}}_2 \left (  \mathbf{X} \right ) +
    \ldots +
    {\beta}_M {\mathbf{h}}_M \left (  \mathbf{X} \right ),
    {{y}_{i}}\in \left\{ -1,+1 \right\},
\end{equation}
where $\pmb{\beta} = {\left [ {\beta}_1,{\beta}_2, \ldots, {\beta}_M  \right]}^\textsf{T} \in \mathbb{R}^{ M \times 1}$ is a coefficient vector corresponding to those $M$ weak classifiers, and the weak classifier with the {\em maximum} coefficient in $\pmb{\beta}$ will be selected.

However, $\hat{\mathbf{y}}$ can not be obtained ideally in practical applications since the data is not always clean. For example, on visual tracking problem, training data is generated automatically and class labels are obtained by the trained classifier. Since the classifier is not perfectly correct, there will be several misclassified training samples in the training set. In order to solve this problem, we split the class labels $\mathbf{y}$ into true labels $\hat{\mathbf{y}}$ and noise labels $\mathbf{e}$ such that $\hat{\mathbf{y}}$ is described as $\hat{\mathbf{y}}= \mathbf{y} - \mathbf{e}$. Thus Equation (\ref{equ:equ9}) is changed to,
\begin{equation}\label{equ:equ10}
    \mathbf{y} = \mathbf{\Phi} \pmb{\beta} + \mathbf{e}.
\end{equation}

In addition, as the predicted labels must be positively related to $\mathbf{y}$, we add a non-negativity constraint to these coefficients. The final formulation is shown in Equation (\ref{equ:equ11}).
\begin{equation}\label{equ:equ11}
    \mathbf{y} = \left  [
                    \begin{matrix}
                    \mathbf{\Phi} & \mathbf{I} & -\mathbf{I}
                    \end{matrix}
                 \right ]
                 \left  [
                    \begin{matrix}
                    \pmb{\beta} \\
                    \mathbf{e}^+ \\
                    \mathbf{e}^-
                    \end{matrix}
                 \right ]
                 = \mathbf{Wc}, \ s.t. \quad \mathbf{c} \ge \mathbf{0},
\end{equation}
where $\mathbf{W} = \left  [
                    \begin{matrix}
                    \mathbf{\Phi} & \mathbf{I} & -\mathbf{I}
                    \end{matrix}
                    \right ] \in \mathbb{R} ^ {L \times \left ( M + 2L\right)}$
and $\mathbf{c} = \left  [
                    \begin{matrix}
                    \pmb{\beta}  \\
                    \mathbf{e}^+  \\
                    \mathbf{e}^-
                    \end{matrix}
                  \right ] \in \mathbb{R} ^ {\left ( M + 2L \right) \times 1}$
is a non-negative coefficient vector.

\subsection{Classifier selection using $\ell_1$ minimization}
Based on the above formulation, we herein want to have a sparse solution to (\ref{equ:equ11}). When we obtain a solution, the $\pmb{{\beta}}^{*}$ should be a sparse vector corresponding to weak classifiers and $\mathbf{e}^+$ and $\mathbf{e}^-$ correspond to the class label noises respectively. In our approach, the weak classifier which has the {\em maximum} value in the sparse vector $\pmb{{\beta}}^{*}$ is selected, {\it i.e.},
\begin{equation}\label{equ:equ12}
    h^{sel} = h^{{m}^{*}}, \text{ where } m^{*} = \underset{\tiny {i}} {argmax} \hspace{0.2cm} {{\beta}_i^{*}}.
\end{equation}

The optimal coefficients $\mathbf{c}^{*}$ can be obtained by solving the following $\ell_0$ optimization objective function. However, the $\ell_0$ minimization is NP-hard problem, thus $\ell_1$ minimization is instead.
\begin{equation}\label{equ:equ14}
    \mathbf{c}^{*} = \underset{\tiny {\mathbf{c}}} {argmax} \hspace{0.2cm} {\left \|  \mathbf{c} \right \|}_1 \quad \ s.t. \quad
    \mathbf{y} = \mathbf{Wc}, \ \mathbf{c} \ge \mathbf{0}
\end{equation}
And Equation (\ref{equ:equ14}) can be changed into Lagrangian optimization problem and solved in polynomial time \cite{19}.
\begin{equation}\label{equ:equ15}
    \mathbf{c}^{*} = \underset{\tiny {\mathbf{c}}} {argmax} \hspace{0.2cm}
     {\left \|   \mathbf{Wc} - \mathbf{y}   \right \|}_2^2 +
     \lambda{\left \|  \mathbf{c} \right \|}_1 \quad \ s.t. \quad
     \ \mathbf{c} \ge \mathbf{0}
\end{equation}

\begin{figure}[htbp]
\centering
\includegraphics[width=0.9\textwidth]{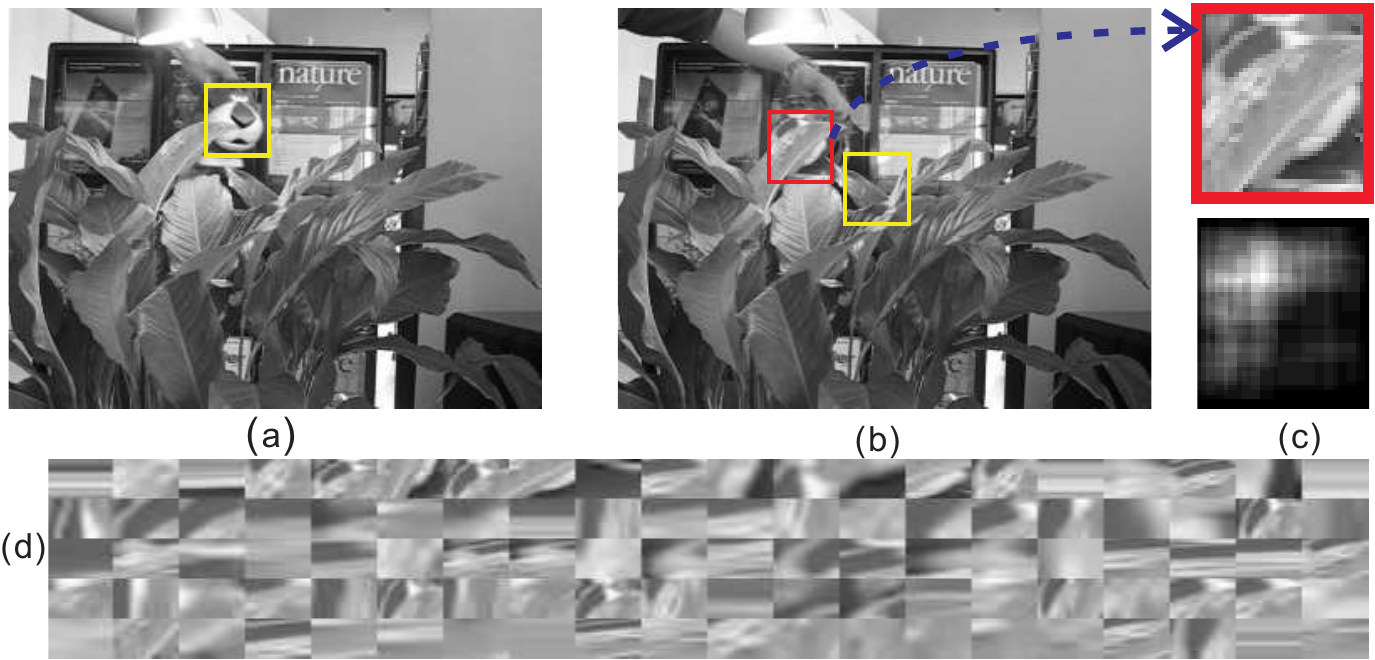}
\caption{Tracking the {\it tiger} with significant pose variations and heavy occlusion. (a) Initial tracking target (yellow rectangle). (b) Tracking results of our method (in red) and expanded online boosting method (in yellow). (c) Spatial distribution of these patches over the red rectangle. (d) Patches corresponding to the classifiers obtained from dynamic classifier selection algorithm. It shows that most of the selected patches are concentrated at the top left part which is not occluded by the tree leaf.}
\label{fig:previewresult}
\end{figure}

Figure~\ref{fig:previewresult} gives some experiments and intermediate results of our method on {\it tiger} sequence.  Figure~\ref{fig:previewresult}(d) shows a sample of patches corresponding to those weak classifiers selected by the proposed algorithm. In order to get the spatial distribution of these patches, we map them to the object rectangle, as shown in Figure~\ref{fig:previewresult}(c).

%===========================================================

\section{Visual tracking using dynamic classifier selection}
\begin{figure}[htbp]
\centering
\includegraphics[width=\textwidth]{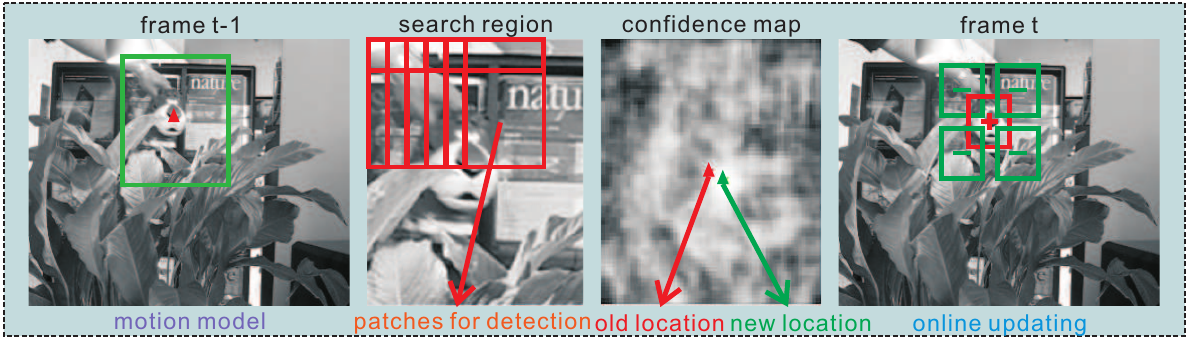}
\caption{
The basic framework of tracking-by-detection algorithms.}
\label{fig:motion}
\end{figure}

In this section we employ the above mentioned dynamic classifier selection algorithm to an expanded online boosting framework. And then a novel tracking approach which is robust to severe occlusion, fast motion, illumination changes and pose variations is proposed. The basic framework of track-by-detection algorithm is depicted in Figure~\ref{fig:motion}. First we will give a brief introduction of motion model. Then the proposed method will be described in detail.

Generally, almost all the motion models in tracking-by-detection algorithms \cite{01,12,13,14,15,16,17} are based on sampling. The goal of tracking is to choose the sample with the {\em maximal} confidence \(p_s \left( y=1| \mathrm{p}_t \right) \) by an online trained classifier, where \(\mathrm{p}_t\) represents the location of estimated 2D patch in the search region at frame \(f_t \in \mathcal{F},t=1,\ldots,T \). As shown in Figure~\ref{fig:motion}, the tracker maintains the object location $\mathrm{p}_{t-1}^{*}$ (marked with red fork) at time $t-1$. In order to estimate the object location at time $t$, a number of patches $\mathrm{P}_t=\left\{ \mathrm{p}_t {\left | \right .} {\left\| \mathrm{p}_t - \mathrm{p}_{t-1}^{*} \right\|}_2^2\le r \right\}$ will be generated to predict the location with a radius $r$. The patch \(\mathrm{p}_t^{*}\) with the {\em maximum} confidence becomes the object location at time $t$.
\begin{equation}\label{equ:equ2}
\mathrm{p}_t^{*}= \underset{\tiny {{ \mathrm{p}_t \in \mathrm{P}_t
}}} {argmax}  \hspace{0.2cm}  p_s \left( y=1 | \mathrm{p}_t
\right)
\end{equation}

This motion model is called translational motion model. Although there exist some more sophisticated motion models, such as particle filter \cite{02,03,10}, combined motion model \cite{06}, we mainly concentrate on the appearance model during tracking, especially in the scheme of online learning.

Given a patch located at \(  \mathrm{p} \in \mathrm{P}_t \), let \( \mathbf{x}^{ \mathrm{p} } \) denote the feature vector extracted from the image patch in the frame  \( f_t \). The training samples are formulated as \( \left \langle {  \mathbf{x}^{ \mathrm{p}}, y } \right \rangle ,y \in \left \{-1,+1\right \} \), where $+1$ and $-1$ represent the sample belonging to the object or background. The strong classifier, \( H  \left ( \mathbf{x}^{  \mathrm{p} } \right) \), has the following form,
\begin{equation}\label{equ:equ3}
    H \left ( \mathbf{x}^{\mathrm{p}} \right ) = \sum\limits_{n=1}^N   \alpha_{n,sel} \cdot h_{n,sel}
    \left ( \mathbf{x}^{\mathrm{p}} \right),
\end{equation}
where \( h_{n,sel} \left ( \mathbf{x}^{\mathrm{p}} \right ) \) is the \(n^{th}\) selected weak classifier and weighted by \(\alpha_{n,sel} \). As the standard online boosting \cite{12} mentioned, \(\alpha_{n,sel}\) can be expressed as $\frac{1}{2} \ln  \frac{1 - e_{n,sel}}{e_{n,sel}} $ where \(e_{n,sel}\) is the accumulated error of the weak classifier and is formulated as $\frac{\lambda_{n,sel}^{w}}{\lambda_{n,sel}^{w} + \lambda_{n,sel}^{c}} $. $\lambda_{n,sel}^{c}$ is the number of samples correctly classified and $\lambda_{n,sel}^{w}$ is the number of samples misclassified.

In order to deal with severe occlusion, we expand online boosting framework by employing part based scheme and propose a part based boosting algorithm in this paper, with the assumption that there will always be a part of the object which is not occluded. The $K$ strong classifiers are combined by adopting the Noisy-OR \cite{27} model. Let \(  \mathrm{p}_t \circ l_k \) indicate the patch location in the \(k^{th}\) part of the object, the model can be written as,

\begin{equation}\label{equ:equ6}
    p_s \left( y=1| \mathrm{p}_t \right) = 1 - \prod_{k=1}^K \left (
    1 - H^k \left ( \mathbf{x}^{ \mathrm{p}_t \circ l_k  }    \right)   \right)
\end{equation}

The algorithm of part based boosting framework and online updating based on dynamic classifier selection are shown in Algorithm 1 and Algorithm 2 respectively.

%===========================================================
%algorithm 1
%===========================================================
\begin{table}[t]
\centering
\vspace{-2mm}
\scalebox{0.93}{
\begin{tabular*}{\textwidth}{l}

  % after \\: \hline or \cline{col1-col2} \cline{col3-col4} ...
%\hline
\noalign{\smallskip}
\hspace{0.0cm}{ {\centering\textbf{Algorithm 1: Part based boosting framework}}}\\
\hline \noalign{\smallskip}
\hspace{0.0cm} {\em\textbf{Input:}}\hspace{0.0cm} Current image $f_t$, object location $\mathrm{p}_{t-1}^{*}$ at time $t-1$.\\
\hspace{0.0cm} {\em\textbf{Output:}}\hspace{0.0cm} Object location $\mathrm{p}_t^{*}$ at time $t$.\\
\hline\noalign{\smallskip}
\hspace{0.0cm} 1 \hspace{0.2cm} $\mathrm{P}_t = \left \{ \mathrm{p}_t  |   {\left \|  \mathrm{p}_t - \mathrm{p}_{t-1}^{*} \right \| }_2^2  \right \}$;\\ \noalign{\smallskip}
\hspace{0.0cm} 2 \hspace{0.2cm} $p_s \left( y=1| \mathrm{p}_t \right) = 1 - \prod_{k=1}^K \left ( 1 - H^k \left ( \mathbf{x}^{ \mathrm{p}_t \circ l_k  }    \right)   \right)$;\\ \noalign{\smallskip}
\hspace{0.0cm} 3 \hspace{0.2cm} $\mathrm{p}_t^{*}= {argmax}_{ \mathrm{p}_t \in \mathrm{P}_t } p_s \left( y=1 | \mathrm{p}_t \right)$;\\ \noalign{\smallskip}
\hspace{0.0cm} 4 \hspace{0.2cm} Generate training samples and label them using the combined strong classifier;\\ \noalign{\smallskip}
\hspace{0.0cm} 5 \hspace{0.2cm} Update each strong classifier using Algorithm 2.\\
\hline
\end{tabular*}
}
\label{alg:alg1}
\end{table}

%===========================================================
%algorithm 2
%===========================================================
\def\Weakc{h_{m,n}^k \left ( \mathbf{x}^{ \mathrm{p}_i \circ l_k } \right )}
\def\TrainS{\left \langle  \mathbf{x}^{ \mathrm{p}_i \circ l_k} , y_i \right \rangle}
\def\WrongC{\lambda_{n,m}^{k,w}}
\def\CorrectC{\lambda_{n,m}^{k,c}}
\def\ErrorC{e_{n,m}^k}
\def\PHI{\mathbf{\Phi}}
\def\MatrixX{\mathbf{X}= \left [
                            \mathbf{x}^{ \mathrm{p}_{1} \circ l_k }, \mathbf{x}^{ \mathrm{p}_2  \circ l_k }, \ldots, \mathbf{x}^{ \mathrm{p}_L \circ l_k}  \right ]}
\def\MatrixY{\mathbf{y}= \left [ y_1, y_2, \ldots, y_L \right ] ^\textsf{T}}
\def\MatrixC{\left [
                    \mathbf{h}_{n,1}^{k} \left (  \mathbf{X} \right ),
                    \mathbf{h}_{n,2}^{k} \left (  \mathbf{X} \right ),
                    \ldots,
                    \mathbf{h}_{n,M}^{k} \left (  \mathbf{X} \right )
                \right]}
\def\MatrixW{\mathbf{W} = \left [
                        \begin{matrix}
                            \mathbf{\Phi} & \mathbf{I} & -\mathbf{I}
                        \end{matrix}
                        \right ]}
\def\VectorC{\mathbf{c}^{*}}
\def\VectorBeta{\pmb{\beta}^{*}}
\def\mStar{m^*}

\begin{table}[t]
\centering
\vspace{-1mm}
\scalebox{0.93}{
\begin{tabular*}{\textwidth}{l}

  % after \\: \hline or \cline{col1-col2} \cline{col3-col4} ...
%\hline
\noalign{\smallskip}
\hspace{0.0cm}{ \textbf{Algorithm 2: Dynamic classifier selection embedded online updating}}\\
\hline \noalign{\smallskip}
\hspace{0.0cm} {\em\textbf{Input:}}\hspace{0.0cm} Training samples $\chi = \left \{
                                 \left \langle  \mathrm{p}_1, y_1  \right \rangle,
                                 \left \langle  \mathrm{p}_2, y_2  \right \rangle,
                                 \ldots,
                                 \left \langle  \mathrm{p}_L, y_L  \right \rangle
                                \right\}$, $y_i \in \left \{ +1, -1 \right\}$  \\
                                \hspace{1.1cm} Strong classifiers $H^k \left ( \mathbf{x} \right ), k = 1,2,\ldots,K$.\\
\hspace{0.0cm} {\em\textbf{Output:}}\hspace{0.0cm} Strong classifiers $H^k \left ( \mathbf{x} \right ), k = 1,2,\ldots,K$.\\
\hline\noalign{\smallskip}
\hspace{0.0cm} 1 \hspace{0.2cm} \textbf{for} $k = 1$ \textbf{to} $K$ \textbf{do}\\ \noalign{\smallskip}
\hspace{0.0cm} 2 \hspace{1.0cm} \textbf{for} $n = 1$ \textbf{to} $N$ \textbf{do}\\ \noalign{\smallskip}
\hspace{0.0cm} 3 \hspace{1.8cm} \textbf{for} $m = 1$ \textbf{to} $M$ \textbf{do}\\ \noalign{\smallskip}
\hspace{0.0cm} 4 \hspace{2.6cm} $\Weakc$ = \tt{Update}$\left ( \Weakc, \TrainS \right )$;\\ \noalign{\smallskip}
\hspace{0.0cm} 5 \hspace{2.6cm} $\WrongC$ = $\WrongC$ + $\sum_i {\mathbf{1} \left ( \Weakc \neq y_i \right)}$;\\ \noalign{\smallskip}
\hspace{0.0cm} 6 \hspace{2.6cm} $\CorrectC$ = $\CorrectC$ + $\sum_i {\mathbf{1} \left ( \Weakc = y_i \right)}$;\\
\hspace{0.0cm} 7 \hspace{2.6cm} $\ErrorC$ = $\frac{\WrongC}{\WrongC + \CorrectC}$;\\
\hspace{0.0cm} 8 \hspace{1.8cm} \textbf{end}\\
\hspace{0.0cm} 9 \hspace{1.8cm} $\MatrixX$, $\MatrixY$;\\ \noalign{\smallskip}
\hspace{0.0cm} 10 \hspace{1.6cm} $\PHI$ = $\MatrixC$, $\MatrixW$;\\ \noalign{\smallskip}
\hspace{0.0cm} 11 \hspace{1.6cm} get $\VectorC$ via solving Equation (\ref{equ:equ15});\\ \noalign{\smallskip}
\hspace{0.0cm} 12 \hspace{1.6cm} $\mStar$ = ${argmax}_i{\VectorBeta}$;\\
\hspace{0.0cm} 13 \hspace{1.6cm} $h_{n, sel}^{k} = h_{n, \mStar}^{k}, \alpha_{n,sel}^k = \frac{1}{2} \ln   \frac{1 - e_{n,\mStar}^{k}}{e_{n,\mStar}^{k}}$;\\
\hspace{0.0cm} 14 \hspace{1.0cm} \textbf{end}\\
\hspace{0.0cm} 15 \hspace{1.0cm} $H^k = \sum\limits_{n=1}^N { \alpha_{n,sel}^{k} \cdot  h_{n,sel}^{k}}$;\\
\hspace{0.0cm} 16 \hspace{0.2cm} \textbf{end}\\
\hline
\end{tabular*}
}
\label{alg:alg2}
\end{table}

\section{Experiments}
In this section, we verify the proposed tracker on 12 challenging tracking sequences ({\it e.g. animal, coke11, david, dollar, faceocc, faceocc2, girl, shaking, surf, sylv, tiger1, tiger2}). The $1^{st}$ and $8^{th}$ sequence can be obtained from \cite{06}, and the others are available at \cite{01}. The proposed method is compared to the latest state-of-the-art tracking algorithms, including fragment-based tracker (FRAG) \cite{11}, $\ell_1$ tracker ($\ell_1$) \cite{02}, multiple instance learning (MIL) tracker \cite{01}, visual tracking decomposition (VTD) approach \cite{06} and the method of part based online boosting without $\ell_1$ minimization called ours (w/o $\ell_1$) using the same initial position. The source codes of FRAG\footnote[1]{\url{http://www.cs.technion.ac.il/~amita/fragtrack/fragtrack.htm}}, $\ell_1$-tracker\footnote[2]{\url{http://www.dabi.temple.edu/~hbling/code_data.htm}}, MIL\footnote[3]{\url{http://vision.ucsd.edu/~bbabenko/project_miltrack.shtml}}, VTD\footnote[4]{\url{http://cv.snu.ac.kr/research/~vtd/}} can be found at URLs.

Our proposed algorithm is implemented in MATLAB, and can process about 15 frames per second. The features we used in this paper are the standard Haar-like features \cite{28}. Each weak classifier $h\left( \mathbf{x} \right)$ corresponds to a Haar-like feature. We use the following weak classifier model,
\begin{equation}\label{equ:equ16}
    h \left (  \mathbf{x}^{\mathrm{p}}\right) =
        \left
        \{
            \begin{array}{c c}
                +1 & \quad p_w \left(  y=1 | \mathbf{x}^{\mathrm{p}} \right) \ge p_w \left(  y=-1 | \mathbf{x}^{\mathrm{p}} \right) \\
                -1 & \quad {otherwise}\\
            \end{array}
        \right.
\end{equation}
where $p_w \left(  y=1 | \mathbf{x}^{\mathrm{p}} \right)$ and $p_w \left(  y=-1 | \mathbf{x}^{\mathrm{p}} \right)$ are two Gaussian distributions, {\em i.e.}, $\mathcal{N}\left( {{\mu }^{+}},{{\sigma }^{+}} \right)$ for positive samples and $\mathcal{N}\left( {{\mu }^{-}},{{\sigma }^{-}} \right)$ for negative samples respectively. In addition, we adopt Kalman filter \cite{09} to update weak classifiers.

Almost all the parameters in the experiments are fixed. We set the number of strong classifiers $K=5$, corresponding to the raw position, half top, half bottom, half left and half right respectively. The number of weak classifier $N$ is $10$ for sequence ({\em coke11, dollar, faceocc, sylv}) and $20$ for others, and the number of weak classifiers $M$ in the pool for selection is $200$. The parameter $\lambda$ in $\ell_1$ minimization is fixed to $0.01$.

In order to eliminate the effects which are brought by randomness, we run each video sequence $5$ times and average the results.

%===========================================================
%===========================================================
%===========================================================
%===========================================================

\subsection{Qualitative Evaluation}
\begin{figure}[tbp]
\centering
\includegraphics[height=0.95\textheight]{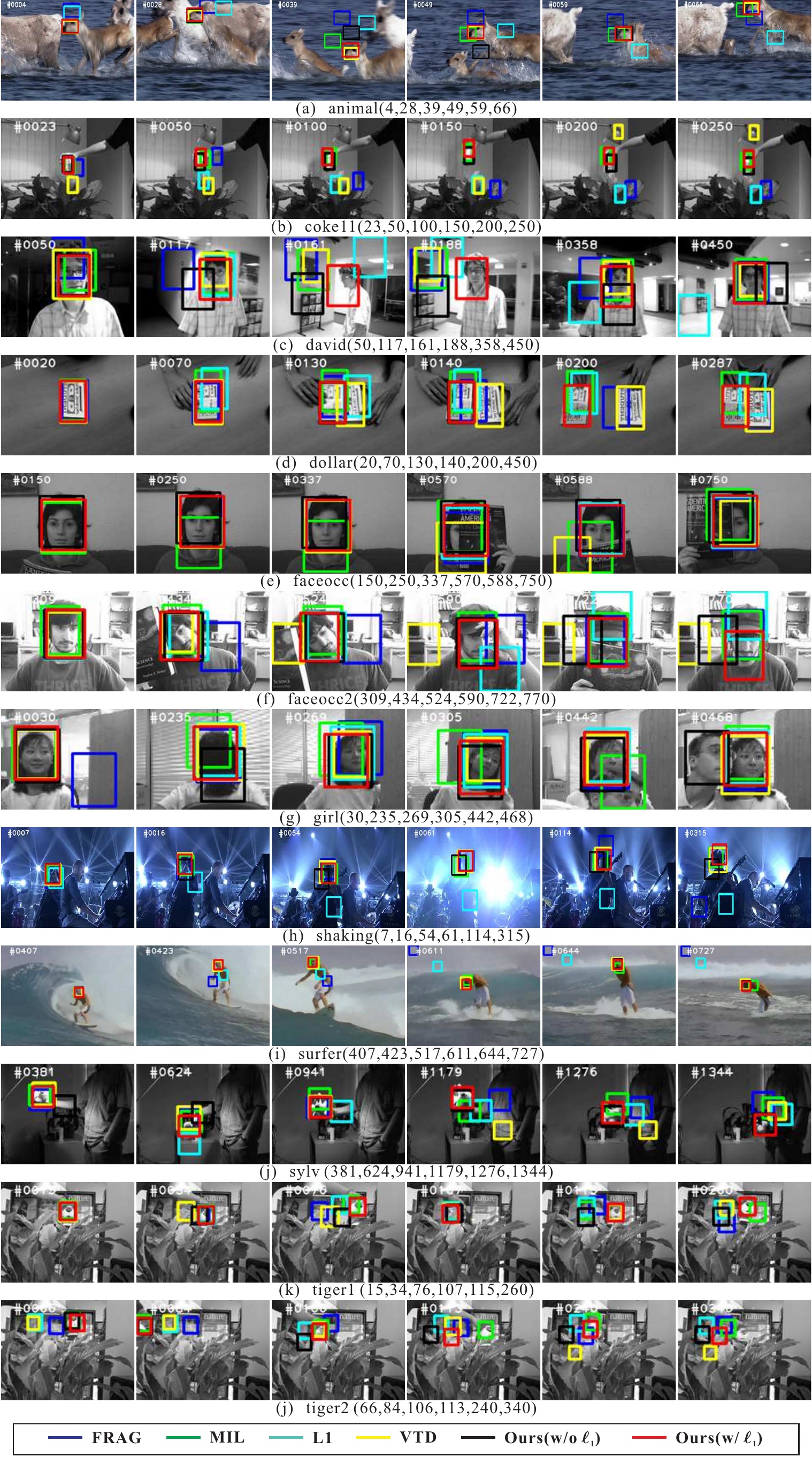}
\caption{
Tracking results of different algorithms for 12 challenging sequences.}
\label{fig:tracking_result}
\end{figure}

In the sequence {\em animal}, several deers are running in water very fast. Figure~\ref{fig:tracking_result}(a) shows the tracking results on {\em animal} sequence. FRAG, MIL tracker, $\ell_1$ tracker, ours (w/o $\ell_1$) tracker failed at frame $39$ because of fast motion. Only ours (w/ $\ell_1$) and VTD tracker can successfully track the object throughout this sequence. The result shows the robustness of ours (w/ $\ell_1$) algorithm in handling fast motion.

In the sequence {\em coke11}, a moving coke can is suffering illumination changes, occlusion and pose variations. The tracking results are shown in Figure~\ref{fig:tracking_result}(b). The FRAG, $\ell_1$ tracker and VTD tracker failed at frame $50$ after illumination variation. The other algorithms provide robust tracking result for this sequence.

In the third sequence {\em david}, light and pose changes significantly throughout the whole sequence. As shown in Figure~\ref{fig:tracking_result}(c), the pose of David changes heavily between frame $117$ and $161$. And all the other algorithms drift when the pose varies wildly. This result clearly demonstrates that ours (w/ $\ell_1$) can handle pose change robustly.

In the sequence {\em dollar}, some dollars are folded and then divided into two parts. The appearance is changed when dollars are folded. From Figure~\ref{fig:tracking_result}(d), we can conclude that the FRAG, $\ell_1$ tracker, and VTD tracker failed to track the dollars when the appearance changed. Ours (w/ $\ell_1$) and (w/o $\ell_1$) and MIL can track the dollars through the whole sequence. The reason for the better performance is that these algorithms are online learning based algorithms.

In the sequence {\em faceocc}, the woman's face is partly occluded by a book. Figure~\ref{fig:tracking_result}(e) shows the tracking results. MIL tracker drifts at frame $250$ after the occlusion. VTD tracker fails to track the face between frame $577$ and frame $600$. The other algorithms can track the occluded face throughout the sequence.

The next the sequence is {\em faceocc2}, which is more challenging than {\em faceocc}. Except the occlusion, the man's appearance is changed (wearing a hat) among the sequence. In Figure~\ref{fig:tracking_result}(f), FRAG and VTD trackers failed when the head is rotated. $\ell_1$ tracker loses the face when the man wears a hat. Ours (w/o $\ell_1$) algorithm tracks the face poorly after frame $707$. However, ours (w/ $\ell_1$) method is able to track the face throughout the whole sequence.

The results of sequence {\em girl} are shown in Figure~\ref{fig:tracking_result}(g). In this sequence, occlusion and pose variations are taken place during the sequence. Ours (w/ $\ell_1$), $\ell_1$ tracker and VTD tracker can successfully track the girl. The other algorithms failed as the pose change and occlusion.

In the sequence {\em shaking} (Figure~\ref{fig:tracking_result}(h)), the variations of illumination and pose are very severe. The results show that all the other algorithms except ours (w/ $\ell_1$) and VTD tracker drift at frame $61$ when light is changed significantly. We show the robustness of our algorithm to severe illumination and pose changes.

The challenges of sequence {\em surfer} are fast motion, pose change and scale change. As shown in Figure~\ref{fig:tracking_result}(i), only FRAG and $\ell_1$ tracker lost the object during tracking. The other algorithms can faithfully track the object through the sequence.

In the sequence {\em sylv}, a moving animal doll is suffering from lighting variations, scale and pose changes. The results are shown in Figure~\ref{fig:tracking_result}(j). $\ell_1$ tracker failed after the frame $624$ and VTD tracker, FRAG also failed after frame $1179$. Only MIL tracker, ours (w/ $\ell_1$) and ours (w/o $\ell_1$) can track the object through the long sequence.

The results of sequence {\em tiger1} are shown in Figure~\ref{fig:tracking_result}(k). Many trackers drift at frame $76$ because of occlusion and fast motion, pose changes. Only ours (w/ $\ell_1$) can deal with these issues well.

In the last sequence {\em tiger2}, trackers except MIL tracker lost the tiger at frame $113$ (Figure~\ref{fig:tracking_result}(l)). Our algorithm can not perform well on this sequence when fast motion and occlusion occurred simultaneously.
%===========================================================
%===========================================================
%===========================================================
%===========================================================
%===========================================================

\subsection{Quantitative Evaluation}

For the quantitative comparison, position error and overlap criterion in PASCAL VOC \cite{29} are employed. They are computed as,
\begin{equation}\label{equ:equ17}
    error \left ( R_G, R_T \right) = {\left \|
        center \left ( R_G  \right) -  center \left ( R_T \right)
                                      \right \|}_2^2
\end{equation}
\begin{equation}\label{equ:equ18}
    overlap \left ( R_G, R_T \right) = \frac
        {area \left ( R_G \cap R_T  \right)}
        {area \left ( R_G \cup R_T  \right)}
\end{equation}

\begin{figure}[tbp]
\centering
\includegraphics[width=0.9\textwidth]{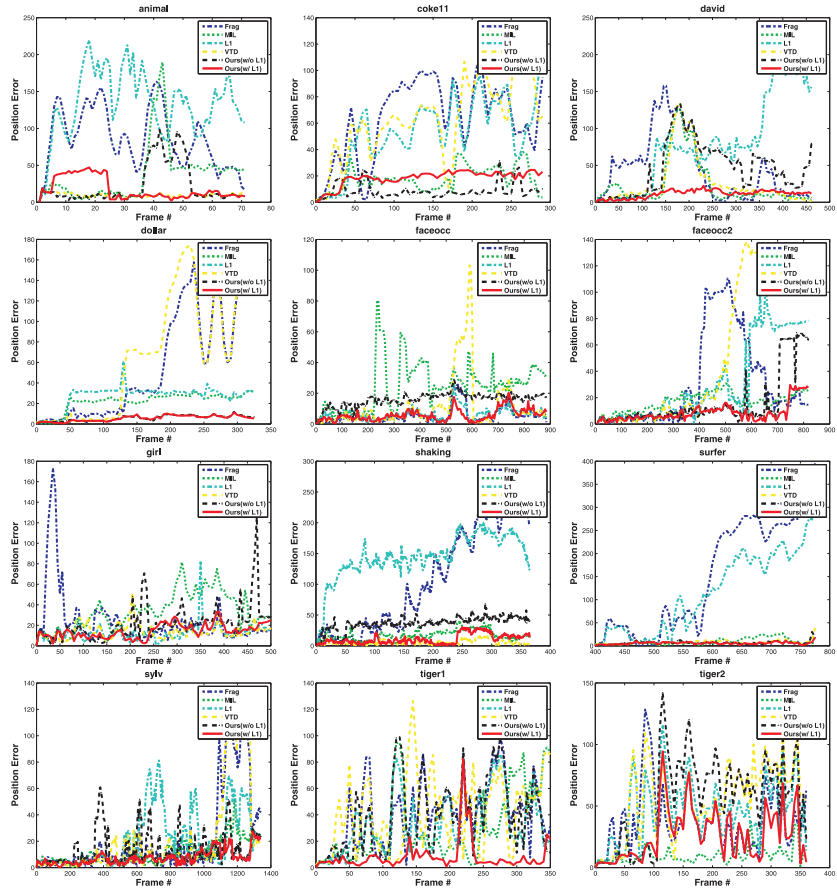}
\caption{Quantitative comparisons of average position error (in pixels).}
\label{fig:position_error}
\end{figure}

Table~\ref{tab:position_error} summarizes the average position error and Table~\ref{tab:success_rate} shows the success rate of tracking methods. Figure~\ref{fig:position_error} depicts the average position error of trackers on each tested sequence. Overall, our proposed algorithm achieves best performance on the sequence {\em david}, {\em dollar}, {\em feaceocc2}, {\em surfer}, {\em sylv}, {\em tiger1} against state-of-the-art methods, and on the other sequences its performance is comparable to the best method.

\section{Conclusion}
In this paper, we propose a dynamic classifier selection algorithm and integrate it to an expanded online boosting framework. In the proposed dynamic classifier selection algorithm, we divided the class labels into two parts which are true and noise class labels. Based on the formulation, the classifier selection is solved by $\ell_1$ {\it minimization}. In the proposed online boosting framework, we employ the part-based idea in which object is represented by several patches. Compared to some state-of-the-art trackers on 12 challenging sequences, the experimental results demonstrate that the proposed method is more accurate and more robust to appearance variations, pose changes, occlusions, fast motion and so on.

\begin{table}[tbp]
%\vspace{-2mm}
\begin{center}
\caption{Average position error of 6 trackers (best: red, second: blue).}
\label{tab:position_error}
\scalebox{0.9}{
    \begin{tabular}{lcccccc}
    \hline\noalign{\smallskip}
    & $\quad$ FRAG $\quad$ & $\quad$ MIL  $\quad$ & $\quad$ $\ell_1$  $\quad$ & $\quad$ VTD  $\quad$ & Ours(w/o $\ell_1$)  & Ours(w/ $\ell_1$)  \\
    \hline\noalign{\smallskip}
    animal    & 86.9  & 37.3  & 133.4  & \color{red}\textbf{9.7} & 23.9  &  \color{blue}\textbf{18.5} \\
    coke11    & 63.3  & 25.4  & 52.7  & 62.8  & \color{red}{\textbf{8.8}} & \color{blue}{\textbf{20.0}}\\
    david     & 45.9  & \color{blue}{\textbf{25.4}} & 78.1  & 26.5  & 46.4  & \color{red}{\textbf{11.5}}\\
    dollar    & 56.2  & 39.3  & 27.1  & 65.7  & \color{blue}{\textbf{5.9}} & \color{red}{\textbf{5.5}} \\
    faceocc   & \color{red}{\textbf{5.3}} & 24.9  & 6.4   & 11.8  & 16.6  & \color{blue}{\textbf{5.9}} \\
    faceocc2  & 30.9  & 22.7  & 31.2  & 52.3  & \color{blue}{\textbf{16.4}} & \color{red}{\textbf{8.5}} \\
    girl      & 23.9  & 33.6  & \color{red}{\textbf{13.1}} & \color{blue}{\textbf{13.9}} & 22.1  & 14.4  \\
    shaking   & 110.2  & 32.1  & 145.1  & \color{red}{\textbf{5.5}} & 36.1  & \color{blue}{\textbf{11.7}} \\
    surfer    & 149.1  & 9.0   & 114.6  & 7.9   & \color{blue}{\textbf{6.6}} & \color{red}{\textbf{5.7}} \\
    sylv      & 21.4  & \color{blue}{\textbf{11.9}} & 21.5  & 21.9  & 14.1  & \color{red}{\textbf{7.6}} \\
    tiger1    & 39.6  & \color{blue}{\textbf{31.2}} & 33.6  & 42.0  & 34.4  & \color{red}{\textbf{8.2}} \\
    tiger2    & 37.6  & \color{red}{\textbf{9.7}} & 46.8  & 54.1  & 53.8  & \color{blue}{\textbf{27.4}} \\
    \hline\noalign{\smallskip}
    Average & 25.2  & 25.2  & 58.6  & 31.2  & \color{blue}{\textbf{23.8}} & \color{red}{\textbf{12.1}} \\
    \hline
\end{tabular}
}
\end{center}
\end{table}

\begin{table}[tbp]
%\vspace{-6mm}
\begin{center}
\caption{Success rate of 6 trackers (best: red, second: blue).}
\label{tab:success_rate}
\scalebox{0.9}{
    \begin{tabular}{lcccccc}
    \hline\noalign{\smallskip}
    $\quad$  & $\quad$ FRAG $\quad$ & $\quad$ MIL  $\quad$ & $\quad$ $\ell_1$  $\quad$ & $\quad$ VTD  $\quad$ & Ours(w/o $\ell_1$)  & Ours(w/ $\ell_1$)  \\
    \hline\noalign{\smallskip}
    animal    & 0.04  & 0.43  & 0.03  & \color{red}{\textbf{0.99}} & 0.73  & \color{blue}{\textbf{0.78}} \\
    coke11    & 0.08  & 0.18  & 0.14  & 0.07  & \color{red}{\textbf{0.68}} & \color{blue}{\textbf{0.32}} \\
    david     & 0.48  & \color{blue}{\textbf{0.61}} & 0.27  & 0.70  & 0.44  & \color{red}{\textbf{0.94}} \\
    dollar    & 0.41  & 0.68  & 0.20  & 0.39  & \color{red}{\textbf{1.00}} & \color{red}{\textbf{1.00}} \\
    faceocc   & \color{red}{\textbf{1.00}} & 0.79  & \color{red}{\textbf{1.00}}  & 0.91  & 0.88  & \color{red}{\textbf{1.00}} \\
    faceocc2  & 0.39  & 0.84  & 0.64  & 0.58  & \color{blue}{\textbf{0.85}} & \color{red}{\textbf{0.99}}  \\
    girl      & 0.83  & 0.50  & \color{red}{\textbf{0.99}} & {0.95} & 0.89  & \color{blue}{\textbf{0.98}}\\
    shaking   & 0.18  & 0.59  & 0.01  & \color{red}{\textbf{1.00}} & 0.48  & \color{blue}{\textbf{0.84}} \\
    surfer    & 0.14  & 0.64  & 0.08  & 0.68  & \color{blue}{\textbf{0.89}} & \color{red}{\textbf{0.92}} \\
    sylv      & \color{blue}{\textbf{0.72}}  & 0.66  & 0.55  & 0.65  & 0.64  & \color{red}{\textbf{0.90}} \\
    tiger1    & 0.20  & \color{blue}{\textbf{0.33}} & 0.20  & 0.23  & 0.29  & \color{red}{\textbf{0.87}} \\
    tiger2    & 0.12  & \color{red}{\textbf{0.66}} & 0.10  & 0.15  & 0.15  & \color{blue}{\textbf{0.38}} \\
    \hline\noalign{\smallskip}
    Average & 0.58  & 0.58  & 0.35  & 0.61  & \color{blue}{\textbf{0.66}} & \color{red}{\textbf{0.83}} \\
    \hline
\end{tabular}
}
\end{center}
\end{table}

\vspace{3mm}
\noindent {\bf Acknowledgement}. This work was supported by NSFC funds (61103060 and 61272287), National ``863''
Programs under grant (2012AA011803) and graduate starting seed fund of Northwestern Polytechnical University (Z2012135), P. R. China.

%===========================================================
\bibliographystyle{splncs}

%this would normally be the end of your paper, but you may also have an appendix
%within the given limit of number of pages
%\end{document}

\end{document}